\documentclass[10pt,letterpaper]{article}

\usepackage{cogsci}

\cogscifinalcopy 

\usepackage{pslatex}
\usepackage{apacite}
\usepackage{float} 

\usepackage{amsmath}
\usepackage{graphicx}
\usepackage{caption}
\usepackage{subcaption}

\DeclareMathOperator{\J}{J}
\DeclareMathOperator{\SB}{S}

\title{Psychophysical-Score: A Behavioral Measure for Assessing the Biological Plausibility of Visual Recognition Models}

\author{
  Brandon RichardWebster\\
  Kitware, Inc.\\
  \And
  Justin Dulay \\
  University of Notre Dame\\
  \AND
  Anthony DiFalco\\
  SPRX\\
   \And
   Walter J. Scheirer\\
   University of Notre Dame
}

\begin{document}
\maketitle

\begin{abstract}
    For the last decade, convolutional neural networks (CNNs) have vastly superseded their predecessors in nearly all vision tasks in artificial intelligence, including object recognition. However, despite abundant advancements, they continue to pale in comparison to biological vision. This chasm has prompted the development of biologically-inspired models that have attempted to mimic the human visual system, primarily at a neural level, which is evaluated using standard dataset benchmarks. However, more work is needed to understand how these models perceive the visual world. This article proposes a state-of-the-art procedure that generates a new metric, Psychophysical-Score, which is grounded in visual psychophysics and is capable of reliably estimating perceptual responses across numerous models --- representing a large range in complexity and biological inspiration. We perform the procedure on twelve models that vary in degree of biological inspiration and complexity, we compare the results against the aggregated results of 2,390 Amazon Mechanical Turk workers who together provided $\sim2.7$ million perceptual responses. Each model's Psychophysical-Score is compared against the state-of-the-art neural activity-based metric, Brain-Score. Our study indicates that models with a high correlation to human perceptual behavior also have a high correlation with the corresponding neural activity.
\end{abstract}

\section{Introduction} \label{sec:intro}
Could biologically-inspired artificial vision --- that is to say, artificial vision possessing neural-like connections or a model of the dynamics of neural activity --- be the next frontier for computer vision scientists? A growing community of neuroscientists, computer vision scientists, funding agencies, and participants in congressional hearings envision this as the next age of artificial vision~\cite{MICrONS51:online,BrainIni97:online,Isneuros47:online,HHRG116A70:online}. Regardless, few experts would be remiss enough to let the current struggles of modern artificial vision systems go unacknowledged. Take convolutional neural networks (CNNs) for example, CNNs require enormous numbers of input examples, easily become overfit, are sensitive to any deviation from the original input, and generally lack robustness to novelty within object classes~\cite{DBLP:journals/corr/abs-1801-00631}. In contrast, a young human child can be shown for the first time a cartoon image of a horse, and with a high degree of accuracy, after just one example, can recognize living horses, pictures of horses, and even cartoon horses --- all of which can differ in breed, color, or artistic style~\cite{guthrie1946psychological}. Human vision requires few examples, generalizes quickly, and is extremely tolerant to many of the conditions that negatively impact CNNs. 


There are two major questions that computer vision scientists need to ask to move forward with the development of more sophisticated artificial object recognition systems that match or exceed human performance: (1) how are models constructed and (2) how does the evaluation of the model affect its construction? Let us begin with the latter. In the task of object recognition in computer vision, for roughly a decade now, progress has been defined and often constrained to rank-1 or rank-5 accuracy on benchmark datasets such as ImageNet~\cite{ILSVRC15} (the one most frequently used), COCO~\cite{lin2014microsoft}, ShapeNet~\cite{shapenet2015}, and Open Images V6~\cite{GoogleAI88:online}. Taken as one pool of data, the combination of images in ImageNet and COCO represents the largest publicly available collection of data for the study of object recognition in computer vision. But it still has the same underlying problem as all datasets. That is, it is a discrete representation of the real world. Thus, while the goal is to improve recognition rates in real-world applications using these dataset(s), they often do not reflect a one-to-one correspondence from the dataset to the real world. If the objective of a computer scientist is to produce models which demonstrate improvement on real-world tasks, then the pursuit of sophisticated object recognition models can be sidetracked by constraining evaluation to just the dataset. 

Coming back to the former point, in the last half-decade, the vast majority of the progress made in computer vision has been made through more complex techniques in deep learning that a) allow for deeper models under the current limits of GPU capacity (with the assumption that deeper models perform better), and b) train more quickly so more input data can be ingested. It is a routine occurrence that with a new generation of GPUs comes a new record performance on traditional metrics. But does the optimization of models on hardware and data equate in the long term to better or more sophisticated artificial object recognition? Not likely. Instead, we need to develop ways to circumvent the constraints of hardware such as small models trained to recognize low-level features like human observers which could improve model generalization.

To help overcome this challenge, we propose a new procedure that generates a new metric, Psychophysical-Score, for examining perceptual similarities between human observers and machine observers. Psychophysical-Score isn't just a metric that helps us quantifiably represent \textit{what} an observer --- whether human or machine ---  perceives but also \textit{why} the observer perceives it that way. Beyond the why, it also could be used to teach a machine observer \textit{how} to perceive like a human observer perceives the \textit{what}. The \textit{how} is particularly possible because Psychophysical-Score doesn't just measure how observers differentiate between objects but also what lies between those objects, the interspace. 

We perform the procedure on seven traditional convolutional neural networks (AlexNet, GoogleNet, ResNet, SqueezeNetv1.0, SqueezeNetv1.1, VGG-16, and VGG-19)~\cite{krizhevsky2014weird,7298594,7780459,iandola2016squeezenet,simonyan2015deep} and five biologically inspired models (Gabor, HMAX, HT-L3, PredNet, and CORnet)~\cite{fogel1989gabor,riesenhuber1999hierarchical,cox2011beyond,lotter-deep-2017,NEURIPS2019-7813d159}, and compare the behavior to 2,390 Amazon Mechanical Turk workers who contributed $\sim2.7$ million perceptual responses. We then perform an analysis that compares Psychophysical-Score results with the published results of the Brain-Score metric~\cite{schrimpf2020integrative}. Our study indicates models which have a high correlation with human perceptual behavior (a high Psychophysical-Score) also have a high correlation with the corresponding neural activity (a high Brain-Score). The positive relationship between neural activity and perceptual behavior indicates a much-needed shift in research from traditional machine learning to machine learning derived from biology --- perceptual behavior, neural activity, or both.

\section{Related Work}
Recent developments between scientists in machine learning, neuroscience, and psychology often use artificial neural networks as a model of the brain. Lotter et al.~\cite{lotter-deep-2017} published PredNet, a biologically inspired predictive coding model which displays similar perceptual behavior to that of human observers on the visual phenomena of the flash-lag effect and end-stopping. Kubilius et al.~\cite{NEURIPS2019-7813d159} introduced a model called CORnet-S, which they report to have similar activation patterns to the neural activity of monkeys. And the most recent addition to artificial neural network models, though not a model by itself, VOneNet~\cite{Dapello2020.06.16.154542}, attempts to model the V1 primary visual cortex of primates. When VOneNet is prepended to the input layer of other artificial neural networks, it is shown to improve protection against white-box adversarial attacks. Each of these recent models represents advancements in both the understanding of biological vision and the quantitative approaches used in artificial neural networks. 

An alternative to expensive and often difficult-to-obtain physiological recordings which were part of these models is a psychological method called psychophysics. Psychophysics is a well-established century-old set of methods for studying the relationship between stimuli and perceptual behavior. In computer vision, psychophysics has been applied to assess whether or not models with claimed high biological fidelity have perceptual behavior that also corresponds to biology. Rajalingham et al.~\cite{rajalingham2018large} compare the recognition behavior of monkeys and people with CNNs, and report that CNNs did not represent the perceptual behavioral patterns of primates. Similar studies have observed the same disparities~\cite{gerhard2013sensitive,heath1996comparison,eberhardt2016deep}. Concerning psychophysics applied to computer vision algorithms specifically, PsyPhy was introduced by RichardWebster et al.~\cite{richardwebster2016psyphy,osti-10144516}. PsyPhy facilitates a psychophysical analysis for object recognition and face recognition through the use of item-response theory~\cite{embretson2000item}. An example finding from RichardWebster et al.~\cite{richardwebster2016psyphy,RichardWebster-2018-ECCV} that is representative of this type of analysis is that there is behavioral consistency between deep learning algorithms and human observers when a blur was applied to input stimuli and in low contrast.

Each of these psychophysical methods used helps us to understand what observers are seeing, but they are lacking in their current form in the capacity to help us understand why. A technique that isn't far off from helping us understand why is the Brain-Score~\cite{schrimpf2020integrative} metric, which is a physiological technique, not a psychophysical technique. Brain-Score is a regularly updated set of benchmarks that compares artificial neural activity with real neural activity derived from the brains of monkeys. CORnet-S gained much of its performance because it incorporated the Brain-Score metric in its design and training process. Though Brain-Score has shown to be vital to biological fidelity, it too suffers from the lack of an answer to the question of why an observer perceives a certain way.

\section{Methods} \label{sec:methods}
\begin{figure}[ht!]
\centering
\begin{subfigure}{0.5\linewidth}
  \centering
  \includegraphics[width=\textwidth]{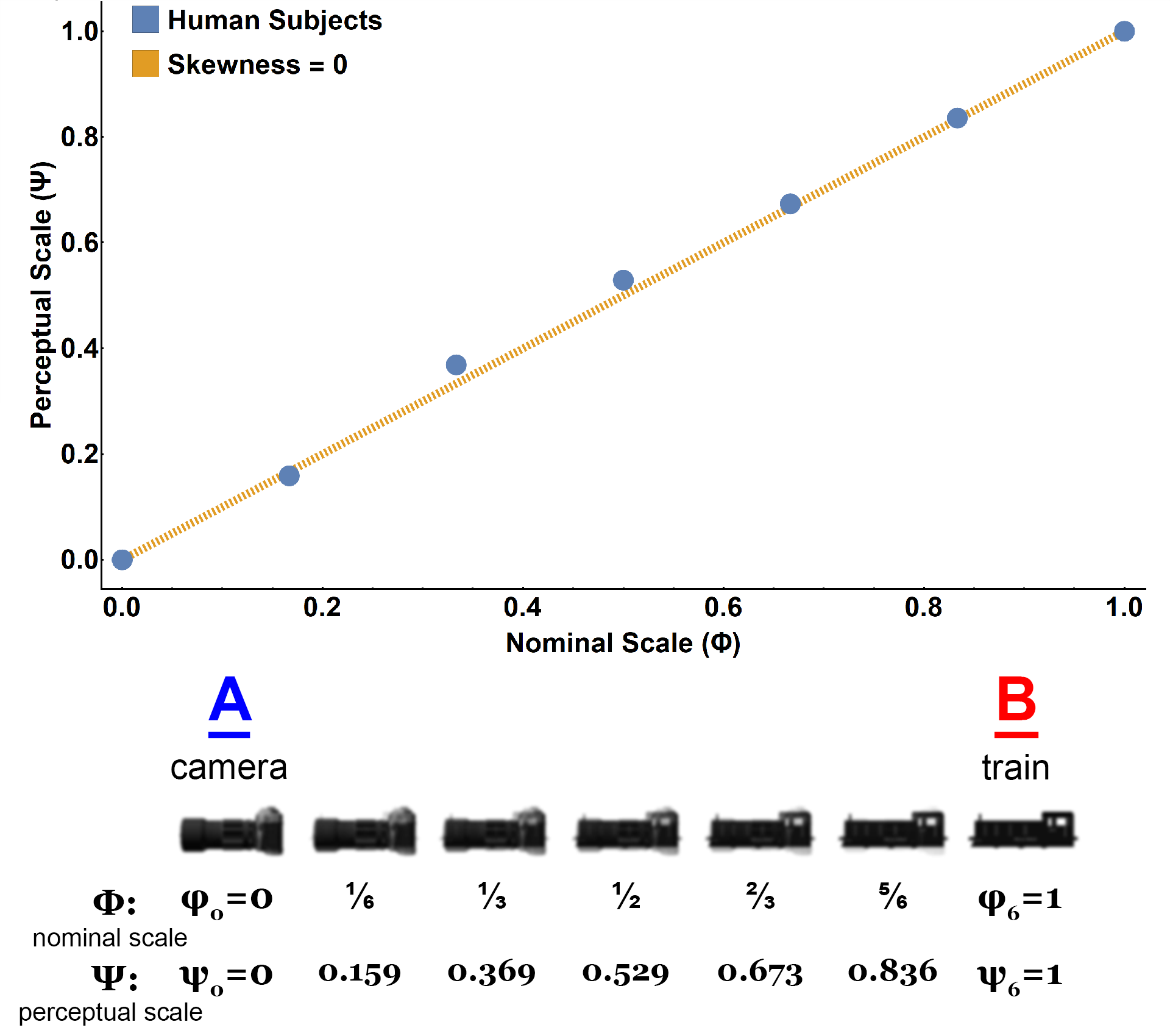}
  \caption{$\Psi$ with no skewness}
  \label{fig:neutral}
\end{subfigure}%
\begin{subfigure}{.5\linewidth}
  \centering
  \includegraphics[width=\textwidth]{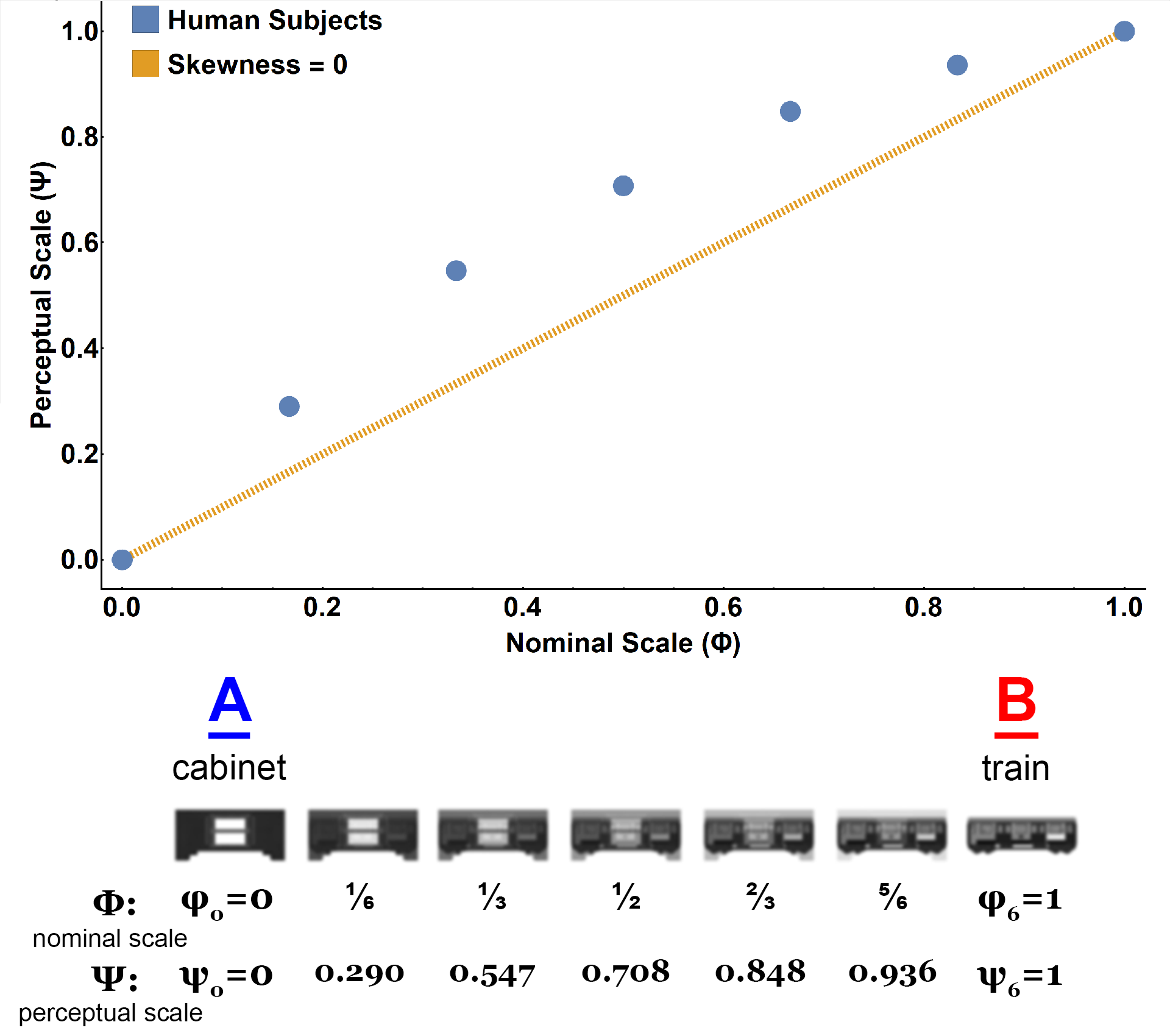}
  \caption{$\Psi$ with positive skewness}
  \label{fig:positive}
\end{subfigure}
\caption{Two plots showing the differences in an unskewed (a) and positively skewed (b) perceptual scale, $\Psi$. Underneath are the object sequences that correspond to the estimated perceptual scale. $\Phi$ is the nominal value/ratio to blend the intermediate objects in the sequence. See Methods Sect. for complete details.}
\label{fig:skewplot}
\end{figure}

\textbf{Maximum-likelihood Difference Scaling (MLDS):}
To understand the interspace between two objects, whether it's the neurological representation in a human observer or the numerical representation in a computer vision observer, we build on a technique called Maximum-likelihood Difference Scaling (MLDS) developed by Maloney et al., 2003~\cite{10.1167/3.8.5}. MLDS is a technique to estimate the parameters of a stochastic model of perceptual differences based on an error measurement. Let us consider Figs.~\ref{fig:neutral} \&~\ref{fig:positive}.

Figs.~\ref{fig:neutral} \&~\ref{fig:positive} each contain a sequence of 7 images where the leftmost image contains an object A, the rightmost object B, and the inner 5 images each contain a composite of A and B. Each image has a corresponding nominal value denoted as $\phi_i$ where $\Phi = \phi_0, \phi_1, ..., \phi_6$ represents the nominal scale for the entire sequence. Recognize that the nominal scale, $\Phi$, is not how observers perceive the differences in the images of the sequence --- although it can be. Figs.~\ref{fig:neutral} \&~\ref{fig:positive}, as will be the case for all sequences presented in this article, has a $\Phi = 0, \frac{1}{6}, ..., 1$. 

While the nominal scale has no direct relationship to an observer's perception, the goal of MLDS is to estimate the parameters of the stochastic model of perceptual differences, $\Psi$. For the remainder of this article, we will either refer to $\Psi$ directly or as a \textit{perceptual scale}. A perceptual scale represents the estimated suprathresholds of perceived distances through the repeated trials in the two-alternative forced choice (2AFC) where an observer is instructed to select the pair (out of two) which is most similar. This choice is repeated numerous times to generate the complete perceptual scale of the sequence. 

\textbf{Stimuli generation:}\label{sec:generation} To generate the stimuli for a perceptual scale we used a software framework called ShapeNet which contains 55 ImageNet classes. There are anywhere from tens to over thousands of instances for each of the 55 classes. There are two ways to define your view-port --- the perspective from which the virtual camera is pointed at an object --- in ShapeNet: panorama and canonical. Panorama allows the user to define the number of steps to take in $360^{\circ}$ of rotation around a selected axis (x, y, z) and direction ($-$ or $+$), and canonical is a selection of six view-ports one for each combination of axis and direction. To simplify our selection and reduce the sheer number of possible generated images (from infinite possibilities), we chose to use the built-in canonical view-port selection.

After we proceeded to generate every object instance with their respective six view-ports, we needed to determine which instance pairs ($a$ and $b$) had adequate overlap before forming a composite of the pair. The higher the overlap in the instance pair, the more significant the object features of the pair will be blended. If the overlap is low, blending of background and object will occur instead and cause the object to appear to fade instead of blending with its pair. To determine which instance pairs we can use for creating composites, we created a mask for each object and computed the Jaccard Index,

\begin{equation}
 \J(a,b) = \frac{|a \cap b|}{|a \cup b|}
\end{equation}

\noindent on the corresponding pair to determine the intersection between $a$ and $b$. The Jaccard Index ranges from $[0,1]$, or no overlap to perfect overlap, respectively. For every instance pair in ShapeNet, we computed the Jaccard index, then selected the 10 pairs --- for each instance pair --- which had the highest Jaccard index. When the Jaccard index was high enough that the top 10 pairs had no statistically significant difference, 10 pairs were randomly selected from that set. Combined with the six canonical perspectives, we started with an initial set of $89.1$k unique pairs. We then generated a sequence for each pair of objects.

Before creating a composite image of $(a,b)$, we needed to ensure each image contained primarily low-level vision features such as edges, gradients, and local morphology.
First, every image is converted to grayscale using the average of the RGB image channels, $\frac{R+B+G}{3}$. Then a Gaussian blur with $\sigma = 3$ is applied to the image to deteriorate higher features. At this point, each image is a grayscale image with each object looking closer to an unrecognizable blob than a known object in the 55 ImageNet classes. 

At this point, each pair was blended using Alpha blending, $A(a^\prime,b^\prime,\alpha) = a^\prime(1-\alpha)+b^\prime(\alpha)$ where $a^\prime$ and $b^\prime$ are the modified source images, and $\alpha$ is in the range $[0,1]$ representing the percentage to include from image $b^\prime$. To generate a sequence, the nominal scale $\Phi = 0, \frac{1}{6}, ..., 1$ is used as input to generate 7 composite images of pair $(a,b)$.

Relating to our terminology, there are two types of sequences: class-level sequences and instance-level sequences. An instance-level sequence is a sequence using pair of images, $(a,b)$, where $a\in A$ and $b \in B$, and $A$ and $B$ are two sets of instances for two classes. An instance-level sequence has 7 composite images but no perceptual scale --- for our experiments --- but a class-sequence has no composite images but has a perceptual scale. A class-level sequence is a set of all instance-level sequences aggregated together using MLDS to create the perceptual scale. In total there were 1450 class-level perceptual scales aggregated. For the remainder of this chapter, we will use the notation $\Psi(A,B)$ to specify the perceptual scale for class $(A,B)$. 


\textbf{Human Observer Experiments:}
A human baseline must be established to analyze how machine learning models compared to their human counterparts. While~\cite{10.1167/3.8.5} used a traditional approach to psychophysical experimentation with in-the-lab monitoring of participants, we use a modern crowd-sourced approach, called Amazon Mechanical Turk (AMT). In all, we had 2,390 participants submitting $\sim2.7$m perceptual responses. Rather than one individual responding to different instances of the same object sequence to obtain the perceptual scale --- as would have been done in a lab setting under the original MLDS paradigm --- we feed the entire set of responses of the participant population into the MLDS optimization routine. Instead of having the set of suprathresholds that might be of one person, we have the mean set of suprathresholds for the population. Due to the vast number of stimuli and needed responses, this was the only way to get the estimations we needed. In this respect, Psychophysical-Score offers a superior approach to Brain-Score because of the difficulty in obtaining large numbers of cortical measurements from subjects.

\textbf{Skewness:}
Perceptual scales appear different for each pair; however, we needed a way to quantitatively compare the 1450 different perceptual scales. We developed a method to measure the skewness of the perceptual space based on the skewness of probability distributions. Recall that the skewness of a probability distribution is a measurement of where the probability mass is located to its tail. For example, in a normal distribution, the probability mass is precisely in the center --- the mean, median, and mode are equal --- with symmetric tails. In a right-tailed positively skewed distribution, the probability mass is closer to zero and the mode $<$ median $<$ mean. A left-tailed negatively skewed distribution has the opposite with mean $>$ median $>$ mode.

Since a perceptual scale has a few similar properties of a probability distribution, we can use the idea of skewness to generate a similar skewness for perceptual scales. However, one fundamental difference is that a perceptual scale is discrete, which poses complications to using previously known skewness calculations. One option to overcome this challenge would be to use a best-fit approximation using a non-symmetric distribution such as the Beta distribution, but this comes at the cost of introducing approximation errors. Alternatively, we can use a discrete integral to approximate the area under the curve and then rescale to match the standard format of skewness $(-\infty,\infty)$:

\begin{equation}
 \SB(\Psi):=-2*\left(\frac{(\sum_{i=0}^{6} \psi_{i}) - 1}{5} - \frac{1}{2}\right)
\end{equation}

\noindent If the $\sum_{}^{}\psi \simeq \frac{1}{2}$, then $\SB(\Psi) \simeq 0$, and is considered to have a neutral interspace. An object sequence with a neutral interspace can be seen in Fig.~\ref{fig:neutral}. If $\sum_{}^{}\psi > \frac{1}{2}$, then $\SB(\Psi) > 0$, which is positive skewness and a positively skewed interspace (see Fig.~\ref{fig:positive}). Finally, if $\sum_{}^{}\psi < \frac{1}{2}$, then $P(\psi) < 0$ has negatively skewed interspace. Given the skewness value for each perceptual scale, we define Psychophysical-Score as the ranked-correlation, monotonic relationship, between a set of perceptual scales measured from humans and those measured from a machine learning model.

\section{Experiments}

\begin{figure}[ht!]
 \centering
    \includegraphics[width=\linewidth,  
]{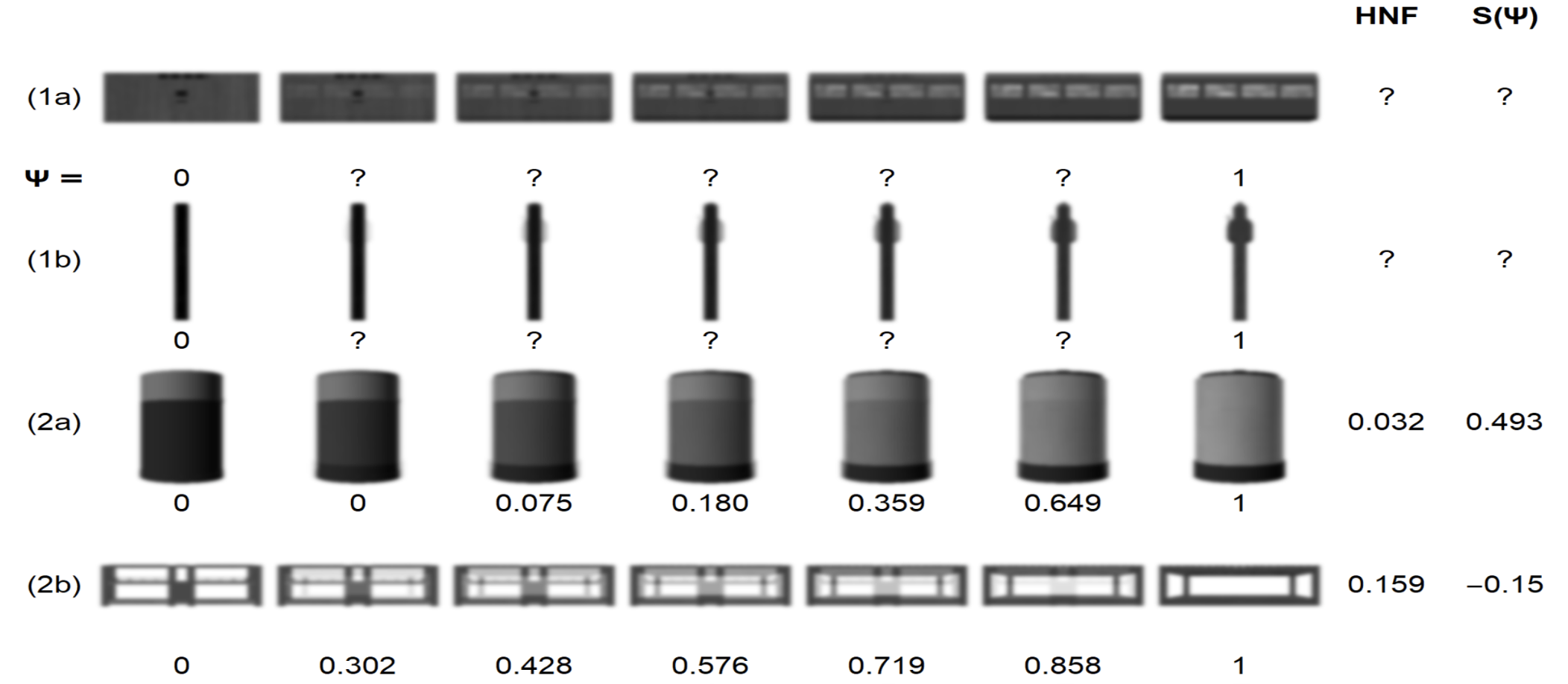}
 \caption{Six object sequences with their corresponding perceptual scale derived from human observers, $\Psi$. The two columns on the right give the value of the human noise factor (HNF) and $\SB(\Psi)$, the skewness of $\Psi$. See Results section for the description of how each sequence 1-4 differs and corresponding analyses.}
 \label{fig:seqs}
\end{figure}

\textbf{What did we learn about human vision?} The original sequences used in Meloney et al.~\cite{10.1167/3.8.5} demonstrated the utility of the MLDS procedure on an ordered sequence of color patches, so the first thing that we needed to determine was if humans could even perform the task with sequences that were significantly more complex (see Fig.~\ref{fig:seqs}). Meloney et al. describe two axiomatic validations an observer must meet for the procedure to work. The first test is that an observer must be able to order the sequence given the complete sequence, and second, meet what is called the Six-Point Property. The first test is a given. The second test is a subset of the first: observers must be able to order any random three images in the sequence (see~\cite{10.1167/3.8.5} for complete details). Out of the 1,485 possible sequences using all 55 objects in ShapeNet, only two sequences did not fully qualify (see Figs.~\ref{fig:seqs}(1a) \&~\ref{fig:seqs}(1b)).

While these two tests indicate observers can properly perform the task for individual sequences, it does not indicate whether the measured suprethresholds for the perceptual space are an adequate representation of the population --- i.e., are the perceptual spaces more than just the product of chance? To determine this, we perform a Pearson's Chi-Squared Test on the null hypothesis, $h_0$, that the perceptual spaces are the product of chance. Each space is represented as a quantitative value, what we call the skewness of the perceptual space (see Methods). The null hypothesis is unequivocally rejected with a $p \cong 0$ and $\alpha < 0.001$. Thus, $h_1$, is true, and the human-obtained perceptual scales are not randomly sampled signals. 

Thus the results so far indicate 1) human observers can perform the task, and 2) the scales that result from the observations mean something beyond the noise. 

\begin{figure}[ht!]
 \centering
    \includegraphics[width=\linewidth,keepaspectratio]{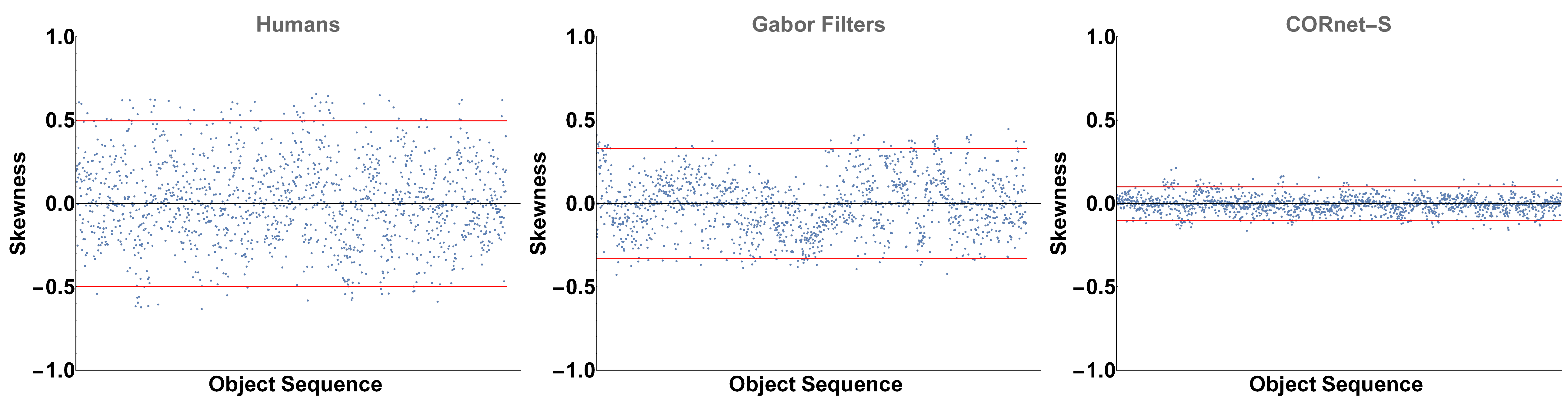}
 \caption{Each skewness, $\SB(\Psi)$, of a perceptual scale, $\Psi$, plotted for three observers: Humans, Gabor, and CORnet-S. The red lines indicate two standard deviations for the observer. Notice the two orders of magnitude difference between the human observers and the CORnet-S observer. Gabor filters have roughly half the amount of variance and have near-random chance level variance. See Table~\ref{tbl:variance} for a complete list of variances by model, and the Results section for further analysis.}
 \label{fig:variance}
\end{figure}

The skewness for each perceptual scale can be seen from a high-level perspective in the ``Human" plot in Fig.~\ref{fig:variance} and Table ~\ref{tbl:variance}. One potential complication in analyzing variance is that AMT presents an uncontrolled environment, which naturally introduces extraneous factors in observer responses. However, even in a controlled environment, there will be variance among observers. To analyze the results of the AMT participants, and consistent with other findings~\cite{germine2011cognitive,10.1371/journal.pone.0057410}, we assume the variance in perception has been accurately captured.

A possible reason for the variance is the varying quality of the object sequences --- not all sequences blend well. For an estimate of the quality of a blend, we can use the human noise factor (HNF) that is estimated along with a perceptual scale using MLDS (see~\cite{10.1167/3.8.5} for complete details). As with perceptual scales, in~\cite{10.1167/3.8.5}, the HNF is for one observer. But for this experiment, it represents the population's noise factor. While we use HNF for blend quality analysis for human perception, the HNF has no bearing on the described procedure to calculate Psychophysical-Score.

\textbf{Human-Machine Learning Comparison:} How do human observers compare to machine learning models? We perform the 2AFC procedure on seven traditional CNNs (AlexNet, GoogleNet, ResNet, SqueezeNetv1.0, SqueezeNetv1.1, VGG-16, and VGG-19)~\cite{krizhevsky2014weird,7298594,7780459,iandola2016squeezenet,simonyan2015deep} and five biologically inspired models (Gabor, HMAX, HT-L3, PredNet, and CORnet)~\cite{fogel1989gabor,riesenhuber1999hierarchical,cox2011beyond,lotter-deep-2017,NEURIPS2019-7813d159}, each representing significant advancements at the time of their publication. The procedure for humans differs from machine observers in two subtle ways: 1) we use many human observers whereas each machine observer is its own, and 2) machine observers use $\ell_{2}$ distance for their difference measurement --- the mechanism by which human observers assess distance is unknown. While procedural differences are minimal, we did make one substantial change to each machine observer that differs from their published form. Before performing the procedure, each machine observer was modified to better match our intentions of measuring low-level vision. In general, the later layers of each model were removed, which are thought to correspond with the later part of the human visual cortex~\cite{dicarlo21}.

Each model performs our proposed procedure so that we can compare scales from human observers to machine observers. Looking at the results in Table~\ref{tbl:variance}, we can see a substantial difference between human and machine performance. Human observers have a larger variance compared to random chance, while all machine observers have variance less than random chance. This is consistent with what one should expect given human perceptual scales are aggregate. However, Gabor filters standout because they are close to random chance, an order of magnitude higher than any other machine observer (Table~\ref{tbl:variance}). Something that differs with Gabor filters that is not present in the other models is that Gabor filters are not trained with data. In this respect, Gabor filters differ the most from human observers as well because humans have learn something visual throughout their lives. Although all of the learned machine models had a variance which was an order of magnitude smaller, it is possible that had each of these models been trained hundreds of times separately to better replicate the crowd-sourcing procedure, the variance would be higher. However, given this and the way the perceptual scales are acquired, we don't believe these fluctuations would be substantial.

\begin{table}[ht!]
\centering
\caption{Ordered List of the variances, $\sigma^2$, in the skewness, $\SB(\Psi)$}
\small
\begin{tabular}{ l r }
\hline
- & $\sigma^2$ \\
Human & 0.06154 \\
Random & 0.03047 \\
Gabor & 0.02704 \\
HT-L3 & 0.00622 \\
HMAX & 0.00607 \\
VGG-19 & 0.00314 \\
AlexNet & 0.00302 \\
SqueezeNet v1.0 & 0.00301 \\
VGG-16 & 0.00264 \\
CORnet-S & 0.00250 \\
GoogleNet & 0.00230 \\
ResNet-18 & 0.00196 \\
SqueezeNet & 0.00133 \\
PredNet & 0.00082 \\
\hline
\end{tabular}
\label{tbl:variance}
\end{table}

In each of the trained machine observers, even the biological ones, the classification layer treats each class as equally important. We don't know yet how human observers comparatively represent objects, but from the results, it is clear that different objects have different representational values. For machine observers, since each object has equal representational value, the skewness of the perceptual space is only present due to the loss in the training regime --- the training regime forces skewness towards zero but only succeeds if there is no error. This is a huge difference between machine and human observers and something the computer vision community needs to consider.

\begin{figure}[ht!]
 \centering
    \includegraphics[width=\linewidth,keepaspectratio]{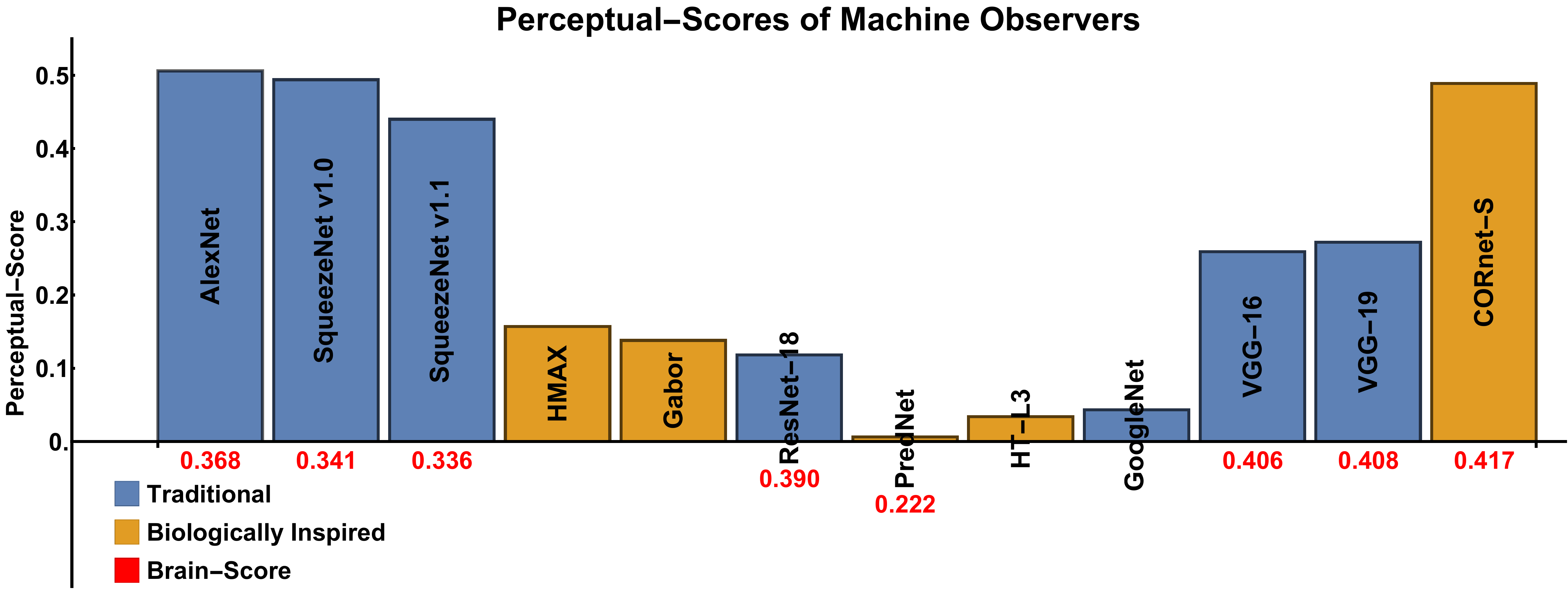}
 \caption{A visualization of the Psychophysical-Scores for each machine observer and each corresponding Brain-Score. Some models do not have a Brain-Score because they are not published on \url{www.brain-score.org}. Note the ``U-shape'' for both Psychophysical-Scores and Brain-Scores, where high Brain-Scores and Psychophysical-Scores are correlated. One outlier exists, ResNet-18. See Results for further analysis of individual Psychophysical-Scores and Brain-Scores for these models.}
 \label{fig:barchart}
\end{figure}

\textbf{Comparison to Brain-Score:} Next we analyze whether or not our procedure produces results that align with the Brain-Score~\cite{schrimpf2020integrative} metric, which is a measurement of a model's biological fidelity from the perspective of neural activity. If a computer vision model has a high Brain-Score, it highly correlates with the neural activity of monkeys, but just because a machine observer has a high correlation, does it mean that the machine observer is ``seeing'' like a human observer? To answer this, we look at the monotonic relationship between the perceptual scales of a machine observer to the perceptual scales of human observers using Spearman's Rho. The correlation of all results can be seen in Fig.~\ref{fig:barchart}. 
Due to the symmetric property of perceptual scales, 
we represent them as either absolute correlation or its Psychophysical-Score. 

The Psychophysical-Score for each machine observer is then compared to its published Brain-Score on \url{www.brain-score.org}. 
When the correlation results of the machine observers are compared as in Fig.~\ref{fig:barchart}, it becomes immediately apparent that Psychophysical-Score and Brain-Score are converging on similar observations, i.e., models which have a higher Psychophysical-Score also have a higher Brain-Score, and vice versa. However, ResNet remains an outlier. A potential reason for this deviation might be that when the later layers were removed, the skip connections which are only present in ResNet, were also removed. Skip connections provide early information to later layers by bypassing intermediate layers. However, the removal of these skip connections may have created a bifurcation in corresponding neural activity (includes the skip connections) and perceptual behavior.

PredNet is a particularly interesting case because it is biologically inspired in design but performs near the bottom on Brain-Score and has ${\sim}0$ as a Psychophysical-Score. PredNet was designed to operationalize a complex predictive coding model of visual processing. This biological mechanism may differ from the neural activity aligned with object recognition, causing a lower Brain-Score. However, this doesn't explain why its Psychophysical-Score is so low. We raise two possibilities for this: 1) by removing all but one predictive block, it may have eliminated quality error propagation (the output before $\ell_{2}$ distance is computed), or 2) behavioral results may differ depending on which of the next predicted frames from the model are used --- we used the first predicted frame.


Finally, and most importantly, the highest scoring model published on \url{www.brain-score.org}, CORnet, also scores high on Psychophysical-Score (although not the highest, there is no statistical difference when compared to AlexNet). CORnet-S is designed for and trained with Brain-Score, which is derived from the neural activity of monkeys, so one would expect CORnet-S to perform high on Brain-Score. But even more, since CORnet-S also has a high Psychophysical-Score, it indicates CORnet-S may have even more similarity to biology than initially considered. First, the neural activity recorded is representing authentic perceptual behavior to some degree. Second, using Brain-Score's representation of neural activity to train a model has significant potential utility. And third, Psychophysical-Score can be used to predict whether a model will have the potential to score high for neural activity, without the need for a difficult and expensive acquisition process --- though the relationship is correlative, not causal.

\section{Discussion}

The capability of human vision has long inspired the pursuit of sophisticated artificial vision systems --- generations of computer vision scientists have worked towards this end. The early advancements in artificial vision were often in tandem with an improved understanding of human visual systems. Unfortunately, the deep learning age has allowed computer vision scientists to lose sight of this original goal of mimicking human visual systems. Thankfully, scientists such as~\cite{schrimpf2020integrative,rajalingham2018large,gerhard2013sensitive,heath1996comparison,eberhardt2016deep,NEURIPS2019-7813d159,lotter-deep-2017,Dapello2020.06.16.154542,richardwebster2016psyphy} have returned to the grassroots of artificial vision. Arguably, the Brain-Score metric published by Schrimpf et al.~\cite{schrimpf2020integrative} is currently the best metric scientists have at estimating the relationship between modern vision algorithms and the neural activity of the brain. Even on its own, Brain-Score has shown to have huge potential~\cite{NEURIPS2019-7813d159}. But now that Psychophysical-Score can complement Brain-Score by estimating the relationship between the perception of machine and human observers, the opportunity for the development of new biologically-inspired algorithms is apparent.

Psychophysical-Score and Brain-Score have the unique ability to inform scientists at every scale of artificial vision algorithm development. Take for example something as simple as the skip connections in ResNet-18. Previously it has been hypothesized that they model the pyramidal neurons in the cerebral cortex. But if Brain-Score and Psychophysical-Score differs so greatly, we should ask questions such as: does this really model a component of the brain? Did we misunderstand that component of the brain? Or is there a missing piece we don't know about? While previously questions like these have been asked rhetorically, they can now be asked with the hope of at least obtaining a partial answer. 
Instead of artificial models learning \textit{what} to perceive from input images, using metrics like Psychophysical-Score and Brain-Score, a model can also learn \textit{how} to perceive directly from human perception and neural activity. If a model can learn how to perceive, maybe it can learn from humans how to do exceedingly difficult tasks such as one-shot learning, generalization, or behavioral prediction.

\bibliographystyle{apacite}

\setlength{\bibleftmargin}{.125in}
\setlength{\bibindent}{-\bibleftmargin}

\bibliography{main}

\begin{thebibliography}{}

\bibitem [\protect \citeauthoryear {%
Chang%
\ \protect \BOthers {.}}{%
Chang%
\ \protect \BOthers {.}}{%
{\protect \APACyear {2015}}%
}]{%
shapenet2015}
\APACinsertmetastar {%
shapenet2015}%
\begin{APACrefauthors}%
Chang, A\BPBI X.%
, Funkhouser, T.%
, Guibas, L.%
, Hanrahan, P.%
, Huang, Q.%
, Li, Z.%
\BDBL {}Yu, F.%
\end{APACrefauthors}%
\unskip\
\newblock
\APACrefYearMonthDay{2015}{}{}.
\newblock
\APACrefbtitle {{ShapeNet: An Information-Rich 3D Model Repository}}
  {{ShapeNet: An Information-Rich 3D Model Repository}}\
  \APACbVolEdTR{}{\BTR{}\ \BNUM\ arXiv:1512.03012 [cs.GR]}.
\newblock
\APACaddressInstitution{}{Stanford University --- Princeton University ---
  Toyota Technological Institute at Chicago}.
\PrintBackRefs{\CurrentBib}

\bibitem [\protect \citeauthoryear {%
Cox%
\ \BBA {} Pinto%
}{%
Cox%
\ \BBA {} Pinto%
}{%
{\protect \APACyear {2011}}%
}]{%
cox2011beyond}
\APACinsertmetastar {%
cox2011beyond}%
\begin{APACrefauthors}%
Cox, D\BPBI D.%
\BCBT {}\ \BBA {} Pinto, N.%
\end{APACrefauthors}%
\unskip\
\newblock
\APACrefYearMonthDay{2011}{}{}.
\newblock
{\BBOQ}\APACrefatitle {Beyond simple features: A large-scale feature search
  approach to unconstrained face recognition} {Beyond simple features: A
  large-scale feature search approach to unconstrained face
  recognition}.{\BBCQ}
\newblock
\BIn{} \APACrefbtitle {IEEE FG.} {Ieee fg.}
\PrintBackRefs{\CurrentBib}

\bibitem [\protect \citeauthoryear {%
Crump%
, McDonnell%
\BCBL {}\ \BBA {} Gureckis%
}{%
Crump%
\ \protect \BOthers {.}}{%
{\protect \APACyear {2013}}%
}]{%
10.1371/journal.pone.0057410}
\APACinsertmetastar {%
10.1371/journal.pone.0057410}%
\begin{APACrefauthors}%
Crump, M\BPBI J\BPBI C.%
, McDonnell, J\BPBI V.%
\BCBL {}\ \BBA {} Gureckis, T\BPBI M.%
\end{APACrefauthors}%
\unskip\
\newblock
\APACrefYearMonthDay{2013}{03}{}.
\newblock
{\BBOQ}\APACrefatitle {Evaluating Amazon's Mechanical Turk as a Tool for
  Experimental Behavioral Research} {Evaluating amazon's mechanical turk as a
  tool for experimental behavioral research}.{\BBCQ}
\newblock
\APACjournalVolNumPages{PLOS ONE}{8}{3}{1-18}.
\newblock
\begin{APACrefURL} \url{https://doi.org/10.1371/journal.pone.0057410}
  \end{APACrefURL}
\newblock
\begin{APACrefDOI} \doi{10.1371/journal.pone.0057410} \end{APACrefDOI}
\PrintBackRefs{\CurrentBib}

\bibitem [\protect \citeauthoryear {%
Dapello%
\ \protect \BOthers {.}}{%
Dapello%
\ \protect \BOthers {.}}{%
{\protect \APACyear {2020}}%
}]{%
Dapello2020.06.16.154542}
\APACinsertmetastar {%
Dapello2020.06.16.154542}%
\begin{APACrefauthors}%
Dapello, J.%
, Marques, T.%
, Schrimpf, M.%
, Geiger, F.%
, Cox, D\BPBI D.%
\BCBL {}\ \BBA {} DiCarlo, J\BPBI J.%
\end{APACrefauthors}%
\unskip\
\newblock
\APACrefYearMonthDay{2020}{}{}.
\newblock
{\BBOQ}\APACrefatitle {Simulating a Primary Visual Cortex at the Front of CNNs
  Improves Robustness to Image Perturbations} {Simulating a primary visual
  cortex at the front of cnns improves robustness to image
  perturbations}.{\BBCQ}
\newblock
\APACjournalVolNumPages{bioRxiv}{}{}{}.
\newblock
\begin{APACrefURL}
  \url{https://www.biorxiv.org/content/early/2020/06/17/2020.06.16.154542}
  \end{APACrefURL}
\newblock
\begin{APACrefDOI} \doi{10.1101/2020.06.16.154542} \end{APACrefDOI}
\PrintBackRefs{\CurrentBib}

\bibitem [\protect \citeauthoryear {%
Dickson%
}{%
Dickson%
}{%
{\protect \APACyear {2021}}%
}]{%
Isneuros47:online}
\APACinsertmetastar {%
Isneuros47:online}%
\begin{APACrefauthors}%
Dickson, B.%
\end{APACrefauthors}%
\unskip\
\newblock
\APACrefYearMonthDay{2021}{}{}.
\newblock
\APACrefbtitle {Is neuroscience the key to protecting AI from adversarial
  attacks? | VentureBeat.} {Is neuroscience the key to protecting ai from
  adversarial attacks? | venturebeat.}
\newblock
\APAChowpublished
  {\url{https://venturebeat.com/2021/01/08/is-neuroscience-the-key-to-protecting-ai-from-adversarial-attacks/}}.
\newblock
\APACrefnote{(Accessed on 01/25/2021)}
\PrintBackRefs{\CurrentBib}

\bibitem [\protect \citeauthoryear {%
Eberhardt%
, Cader%
\BCBL {}\ \BBA {} Serre%
}{%
Eberhardt%
\ \protect \BOthers {.}}{%
{\protect \APACyear {2016}}%
}]{%
eberhardt2016deep}
\APACinsertmetastar {%
eberhardt2016deep}%
\begin{APACrefauthors}%
Eberhardt, S.%
, Cader, J.%
\BCBL {}\ \BBA {} Serre, T.%
\end{APACrefauthors}%
\unskip\
\newblock
\APACrefYearMonthDay{2016}{}{}.
\newblock
{\BBOQ}\APACrefatitle {How Deep is the Feature Analysis underlying Rapid Visual
  Categorization?} {How deep is the feature analysis underlying rapid visual
  categorization?}{\BBCQ}
\newblock
\BIn{} \APACrefbtitle {NIPS.} {Nips.}
\PrintBackRefs{\CurrentBib}

\bibitem [\protect \citeauthoryear {%
Embretson%
\ \BBA {} Reise%
}{%
Embretson%
\ \BBA {} Reise%
}{%
{\protect \APACyear {2000}}%
}]{%
embretson2000item}
\APACinsertmetastar {%
embretson2000item}%
\begin{APACrefauthors}%
Embretson, S\BPBI E.%
\BCBT {}\ \BBA {} Reise, S\BPBI P.%
\end{APACrefauthors}%
\unskip\
\newblock
\APACrefYear{2000}.
\newblock
\APACrefbtitle {Item response theory for psychologists} {Item response theory
  for psychologists}.
\newblock
\APACaddressPublisher{}{Lawrence Erlbaum Associates, Inc.}
\PrintBackRefs{\CurrentBib}

\bibitem [\protect \citeauthoryear {%
Fogel%
\ \BBA {} Sagi%
}{%
Fogel%
\ \BBA {} Sagi%
}{%
{\protect \APACyear {1989}}%
}]{%
fogel1989gabor}
\APACinsertmetastar {%
fogel1989gabor}%
\begin{APACrefauthors}%
Fogel, I.%
\BCBT {}\ \BBA {} Sagi, D.%
\end{APACrefauthors}%
\unskip\
\newblock
\APACrefYearMonthDay{1989}{}{}.
\newblock
{\BBOQ}\APACrefatitle {Gabor filters as texture discriminator} {Gabor filters
  as texture discriminator}.{\BBCQ}
\newblock
\APACjournalVolNumPages{Biological cybernetics}{61}{2}{103--113}.
\PrintBackRefs{\CurrentBib}

\bibitem [\protect \citeauthoryear {%
Gerhard%
, Wichmann%
\BCBL {}\ \BBA {} Bethge%
}{%
Gerhard%
\ \protect \BOthers {.}}{%
{\protect \APACyear {2013}}%
}]{%
gerhard2013sensitive}
\APACinsertmetastar {%
gerhard2013sensitive}%
\begin{APACrefauthors}%
Gerhard, H\BPBI E.%
, Wichmann, F\BPBI A.%
\BCBL {}\ \BBA {} Bethge, M.%
\end{APACrefauthors}%
\unskip\
\newblock
\APACrefYearMonthDay{2013}{}{}.
\newblock
{\BBOQ}\APACrefatitle {How sensitive is the human visual system to the local
  statistics of natural images?} {How sensitive is the human visual system to
  the local statistics of natural images?}{\BBCQ}
\newblock
\APACjournalVolNumPages{PLoS Computational Biology}{9}{1}{e1002873}.
\PrintBackRefs{\CurrentBib}

\bibitem [\protect \citeauthoryear {%
Germine%
, Duchaine%
\BCBL {}\ \BBA {} Nakayama%
}{%
Germine%
\ \protect \BOthers {.}}{%
{\protect \APACyear {2011}}%
}]{%
germine2011cognitive}
\APACinsertmetastar {%
germine2011cognitive}%
\begin{APACrefauthors}%
Germine, L\BPBI T.%
, Duchaine, B.%
\BCBL {}\ \BBA {} Nakayama, K.%
\end{APACrefauthors}%
\unskip\
\newblock
\APACrefYearMonthDay{2011}{}{}.
\newblock
{\BBOQ}\APACrefatitle {Where cognitive development and aging meet: Face
  learning ability peaks after age 30} {Where cognitive development and aging
  meet: Face learning ability peaks after age 30}.{\BBCQ}
\newblock
\APACjournalVolNumPages{Cognition}{118}{2}{201--210}.
\PrintBackRefs{\CurrentBib}

\bibitem [\protect \citeauthoryear {%
Guthrie%
}{%
Guthrie%
}{%
{\protect \APACyear {1946}}%
}]{%
guthrie1946psychological}
\APACinsertmetastar {%
guthrie1946psychological}%
\begin{APACrefauthors}%
Guthrie, E\BPBI R.%
\end{APACrefauthors}%
\unskip\
\newblock
\APACrefYearMonthDay{1946}{}{}.
\newblock
{\BBOQ}\APACrefatitle {Psychological facts and psychological theory.}
  {Psychological facts and psychological theory.}{\BBCQ}
\newblock
\APACjournalVolNumPages{Psychological Bulletin}{43}{1}{1}.
\PrintBackRefs{\CurrentBib}

\bibitem [\protect \citeauthoryear {%
{He}%
, {Zhang}%
, {Ren}%
\BCBL {}\ \BBA {} {Sun}%
}{%
{He}%
\ \protect \BOthers {.}}{%
{\protect \APACyear {2016}}%
}]{%
7780459}
\APACinsertmetastar {%
7780459}%
\begin{APACrefauthors}%
{He}, K.%
, {Zhang}, X.%
, {Ren}, S.%
\BCBL {}\ \BBA {} {Sun}, J.%
\end{APACrefauthors}%
\unskip\
\newblock
\APACrefYearMonthDay{2016}{}{}.
\newblock
{\BBOQ}\APACrefatitle {Deep Residual Learning for Image Recognition} {Deep
  residual learning for image recognition}.{\BBCQ}
\newblock
\BIn{} \APACrefbtitle {2016 IEEE Conference on Computer Vision and Pattern
  Recognition (CVPR)} {2016 ieee conference on computer vision and pattern
  recognition (cvpr)}\ (\BPG~770-778).
\newblock
\begin{APACrefDOI} \doi{10.1109/CVPR.2016.90} \end{APACrefDOI}
\PrintBackRefs{\CurrentBib}

\bibitem [\protect \citeauthoryear {%
Heath%
, Sarkar%
, Sanocki%
\BCBL {}\ \BBA {} Bowyer%
}{%
Heath%
\ \protect \BOthers {.}}{%
{\protect \APACyear {1996}}%
}]{%
heath1996comparison}
\APACinsertmetastar {%
heath1996comparison}%
\begin{APACrefauthors}%
Heath, M.%
, Sarkar, S.%
, Sanocki, T.%
\BCBL {}\ \BBA {} Bowyer, K.%
\end{APACrefauthors}%
\unskip\
\newblock
\APACrefYearMonthDay{1996}{}{}.
\newblock
{\BBOQ}\APACrefatitle {Comparison of edge detectors: a methodology and initial
  study} {Comparison of edge detectors: a methodology and initial
  study}.{\BBCQ}
\newblock
\BIn{} \APACrefbtitle {Computer Vision and Pattern Recognition, 1996.
  Proceedings CVPR'96, 1996 IEEE Computer Society Conference on} {Computer
  vision and pattern recognition, 1996. proceedings cvpr'96, 1996 ieee computer
  society conference on}\ (\BPGS\ 143--148).
\PrintBackRefs{\CurrentBib}

\bibitem [\protect \citeauthoryear {%
Iandola%
\ \protect \BOthers {.}}{%
Iandola%
\ \protect \BOthers {.}}{%
{\protect \APACyear {2016}}%
}]{%
iandola2016squeezenet}
\APACinsertmetastar {%
iandola2016squeezenet}%
\begin{APACrefauthors}%
Iandola, F\BPBI N.%
, Han, S.%
, Moskewicz, M\BPBI W.%
, Ashraf, K.%
, Dally, W\BPBI J.%
\BCBL {}\ \BBA {} Keutzer, K.%
\end{APACrefauthors}%
\unskip\
\newblock
\APACrefYearMonthDay{2016}{}{}.
\newblock
{\BBOQ}\APACrefatitle {SqueezeNet: AlexNet-level accuracy with 50x fewer
  parameters and< 0.5 MB model size} {Squeezenet: Alexnet-level accuracy with
  50x fewer parameters and< 0.5 mb model size}.{\BBCQ}
\newblock
\APACjournalVolNumPages{arXiv preprint arXiv:1602.07360}{}{}{}.
\PrintBackRefs{\CurrentBib}

\bibitem [\protect \citeauthoryear {%
IARPA%
}{%
IARPA%
}{%
{\protect \APACyear {{\protect \bibnodate {}}}}%
}]{%
MICrONS51:online}
\APACinsertmetastar {%
MICrONS51:online}%
\begin{APACrefauthors}%
IARPA, I\BPBI A\BPBI R\BPBI P\BPBI A.%
\end{APACrefauthors}%
\unskip\
\newblock
\APACrefYearMonthDay{{\protect \bibnodate {}}}{}{}.
\newblock
\APACrefbtitle {MICrONS.} {Microns.}
\newblock
\APAChowpublished
  {\url{https://www.iarpa.gov/index.php/research-programs/microns}}.
\newblock
\APACrefnote{(Accessed on 01/25/2021)}
\PrintBackRefs{\CurrentBib}

\bibitem [\protect \citeauthoryear {%
Kasthuri%
}{%
Kasthuri%
}{%
{\protect \APACyear {2020}}%
}]{%
HHRG116A70:online}
\APACinsertmetastar {%
HHRG116A70:online}%
\begin{APACrefauthors}%
Kasthuri, N.%
\end{APACrefauthors}%
\unskip\
\newblock
\APACrefYearMonthDay{2020}{}{}.
\newblock
\APACrefbtitle {HHRG-116-AP10-Wstate-KasthuriN-20200205.pdf.}
  {Hhrg-116-ap10-wstate-kasthurin-20200205.pdf.}
\newblock
\APAChowpublished
  {\url{https://docs.house.gov/meetings/AP/AP10/20200205/110447/HHRG-116-AP10-Wstate-KasthuriN-20200205.pdf}}.
\newblock
\APACrefnote{(Accessed on 01/25/2021)}
\PrintBackRefs{\CurrentBib}

\bibitem [\protect \citeauthoryear {%
Krizhevsky%
}{%
Krizhevsky%
}{%
{\protect \APACyear {2014}}%
}]{%
krizhevsky2014weird}
\APACinsertmetastar {%
krizhevsky2014weird}%
\begin{APACrefauthors}%
Krizhevsky, A.%
\end{APACrefauthors}%
\unskip\
\newblock
\APACrefYearMonthDay{2014}{}{}.
\newblock
\APACrefbtitle {One weird trick for parallelizing convolutional neural
  networks.} {One weird trick for parallelizing convolutional neural networks.}
\PrintBackRefs{\CurrentBib}

\bibitem [\protect \citeauthoryear {%
Kubilius%
\ \protect \BOthers {.}}{%
Kubilius%
\ \protect \BOthers {.}}{%
{\protect \APACyear {2019}}%
}]{%
NEURIPS2019-7813d159}
\APACinsertmetastar {%
NEURIPS2019-7813d159}%
\begin{APACrefauthors}%
Kubilius, J.%
, Schrimpf, M.%
, Kar, K.%
, Rajalingham, R.%
, Hong, H.%
, Majaj, N.%
\BDBL {}DiCarlo, J\BPBI J.%
\end{APACrefauthors}%
\unskip\
\newblock
\APACrefYearMonthDay{2019}{}{}.
\newblock
{\BBOQ}\APACrefatitle {Brain-Like Object Recognition with High-Performing
  Shallow Recurrent ANNs} {Brain-like object recognition with high-performing
  shallow recurrent anns}.{\BBCQ}
\newblock
\BIn{} \APACrefbtitle {Advances in Neural Information Processing Systems}
  {Advances in neural information processing systems}\ (\BVOL~32, \BPGS\
  12805--12816).
\newblock
\APACaddressPublisher{}{Curran Associates, Inc.}
\newblock
\begin{APACrefURL}
  \url{https://proceedings.neurips.cc/paper/2019/file/7813d1590d28a7dd372ad54b5d29d033-Paper.pdf}
  \end{APACrefURL}
\PrintBackRefs{\CurrentBib}

\bibitem [\protect \citeauthoryear {%
Lin%
\ \protect \BOthers {.}}{%
Lin%
\ \protect \BOthers {.}}{%
{\protect \APACyear {2014}}%
}]{%
lin2014microsoft}
\APACinsertmetastar {%
lin2014microsoft}%
\begin{APACrefauthors}%
Lin, T\BHBI Y.%
, Maire, M.%
, Belongie, S.%
, Hays, J.%
, Perona, P.%
, Ramanan, D.%
\BDBL {}Zitnick, C\BPBI L.%
\end{APACrefauthors}%
\unskip\
\newblock
\APACrefYearMonthDay{2014}{}{}.
\newblock
{\BBOQ}\APACrefatitle {Microsoft coco: Common objects in context} {Microsoft
  coco: Common objects in context}.{\BBCQ}
\newblock
\BIn{} \APACrefbtitle {European conference on computer vision} {European
  conference on computer vision}\ (\BPGS\ 740--755).
\PrintBackRefs{\CurrentBib}

\bibitem [\protect \citeauthoryear {%
Lotter%
, Kreiman%
\BCBL {}\ \BBA {} Cox%
}{%
Lotter%
\ \protect \BOthers {.}}{%
{\protect \APACyear {2017}}%
}]{%
lotter-deep-2017}
\APACinsertmetastar {%
lotter-deep-2017}%
\begin{APACrefauthors}%
Lotter, W.%
, Kreiman, G.%
\BCBL {}\ \BBA {} Cox, D.%
\end{APACrefauthors}%
\unskip\
\newblock
\APACrefYearMonthDay{2017}{}{}.
\newblock
{\BBOQ}\APACrefatitle {Deep predictive coding networks for video prediction and
  unsupervised learning} {Deep predictive coding networks for video prediction
  and unsupervised learning}.{\BBCQ}
\newblock
\BIn{} \APACrefbtitle {Proceedings of the {International} {Conference} on
  {Learning} {Representations}.} {Proceedings of the {International}
  {Conference} on {Learning} {Representations}.}
\PrintBackRefs{\CurrentBib}

\bibitem [\protect \citeauthoryear {%
Maloney%
\ \BBA {} Yang%
}{%
Maloney%
\ \BBA {} Yang%
}{%
{\protect \APACyear {2003}}%
}]{%
10.1167/3.8.5}
\APACinsertmetastar {%
10.1167/3.8.5}%
\begin{APACrefauthors}%
Maloney, L\BPBI T.%
\BCBT {}\ \BBA {} Yang, J\BPBI N.%
\end{APACrefauthors}%
\unskip\
\newblock
\APACrefYearMonthDay{2003}{10}{}.
\newblock
{\BBOQ}\APACrefatitle {{Maximum likelihood difference scaling}} {{Maximum
  likelihood difference scaling}}.{\BBCQ}
\newblock
\APACjournalVolNumPages{Journal of Vision}{3}{8}{5-5}.
\newblock
\begin{APACrefURL} \url{https://doi.org/10.1167/3.8.5} \end{APACrefURL}
\newblock
\begin{APACrefDOI} \doi{10.1167/3.8.5} \end{APACrefDOI}
\PrintBackRefs{\CurrentBib}

\bibitem [\protect \citeauthoryear {%
Marcus%
}{%
Marcus%
}{%
{\protect \APACyear {2018}}%
}]{%
DBLP:journals/corr/abs-1801-00631}
\APACinsertmetastar {%
DBLP:journals/corr/abs-1801-00631}%
\begin{APACrefauthors}%
Marcus, G.%
\end{APACrefauthors}%
\unskip\
\newblock
\APACrefYearMonthDay{2018}{}{}.
\newblock
{\BBOQ}\APACrefatitle {Deep Learning: {A} Critical Appraisal} {Deep learning:
  {A} critical appraisal}.{\BBCQ}
\newblock
\APACjournalVolNumPages{CoRR}{abs/1801.00631}{}{}.
\newblock
\begin{APACrefURL} \url{http://arxiv.org/abs/1801.00631} \end{APACrefURL}
\PrintBackRefs{\CurrentBib}

\bibitem [\protect \citeauthoryear {%
of Health:~NIH%
}{%
of Health:~NIH%
}{%
{\protect \APACyear {2021}}%
}]{%
BrainIni97:online}
\APACinsertmetastar {%
BrainIni97:online}%
\begin{APACrefauthors}%
of Health:~NIH, N\BPBI I.%
\end{APACrefauthors}%
\unskip\
\newblock
\APACrefYearMonthDay{2021}{}{}.
\newblock
\APACrefbtitle {Brain Initiative.} {Brain initiative.}
\newblock
\APAChowpublished {\url{https://braininitiative.nih.gov/}}.
\newblock
\APACrefnote{(Accessed on 01/25/2021)}
\PrintBackRefs{\CurrentBib}

\bibitem [\protect \citeauthoryear {%
Pont-Tuset%
}{%
Pont-Tuset%
}{%
{\protect \APACyear {2020}}%
}]{%
GoogleAI88:online}
\APACinsertmetastar {%
GoogleAI88:online}%
\begin{APACrefauthors}%
Pont-Tuset, J.%
\end{APACrefauthors}%
\unskip\
\newblock
\APACrefYearMonthDay{2020}{}{}.
\newblock
\APACrefbtitle {Google AI Blog: Open Images V6 — Now Featuring Localized
  Narratives.} {Google ai blog: Open images v6 — now featuring localized
  narratives.}
\newblock
\APAChowpublished
  {\url{https://ai.googleblog.com/2020/02/open-images-v6-now-featuring-localized.html}}.
\newblock
\APACrefnote{(Accessed on 01/25/2021)}
\PrintBackRefs{\CurrentBib}

\bibitem [\protect \citeauthoryear {%
Rajalingham%
\ \protect \BOthers {.}}{%
Rajalingham%
\ \protect \BOthers {.}}{%
{\protect \APACyear {2018}}%
{\protect \APACexlab {{\protect \BCnt {1}}}}}]{%
rajalingham2018large}
\APACinsertmetastar {%
rajalingham2018large}%
\begin{APACrefauthors}%
Rajalingham, R.%
, Issa, E\BPBI B.%
, Bashivan, P.%
, Kar, K.%
, Schmidt, K.%
\BCBL {}\ \BBA {} DiCarlo, J\BPBI J.%
\end{APACrefauthors}%
\unskip\
\newblock
\APACrefYearMonthDay{2018{\protect \BCnt {1}}}{}{}.
\newblock
{\BBOQ}\APACrefatitle {Large-scale, high-resolution comparison of the core
  visual object recognition behavior of humans, monkeys, and state-of-the-art
  deep artificial neural networks} {Large-scale, high-resolution comparison of
  the core visual object recognition behavior of humans, monkeys, and
  state-of-the-art deep artificial neural networks}.{\BBCQ}
\newblock
\APACjournalVolNumPages{bioRxiv}{}{}{240614}.
\PrintBackRefs{\CurrentBib}

\bibitem [\protect \citeauthoryear {%
Rajalingham%
\ \protect \BOthers {.}}{%
Rajalingham%
\ \protect \BOthers {.}}{%
{\protect \APACyear {2018}}%
{\protect \APACexlab {{\protect \BCnt {2}}}}}]{%
dicarlo21}
\APACinsertmetastar {%
dicarlo21}%
\begin{APACrefauthors}%
Rajalingham, R.%
, Issa, E\BPBI B.%
, Bashivan, P.%
, Kar, K.%
, Schmidt, K.%
\BCBL {}\ \BBA {} DiCarlo, J\BPBI J.%
\end{APACrefauthors}%
\unskip\
\newblock
\APACrefYearMonthDay{2018{\protect \BCnt {2}}}{02/2018}{}.
\newblock
{\BBOQ}\APACrefatitle {Large-scale, high-resolution comparison of the core
  visual object recognition behavior of humans, monkeys, and state-of-the-art
  deep artificial neural networks} {Large-scale, high-resolution comparison of
  the core visual object recognition behavior of humans, monkeys, and
  state-of-the-art deep artificial neural networks}{\BBCQ}\ [preprint].
\newblock
\APACjournalVolNumPages{bioRxiv}{}{}{}.
\newblock
\begin{APACrefURL}
  \url{https://www.biorxiv.org/content/10.1101/240614v4.full.pdf}
  \end{APACrefURL}
\newblock
\begin{APACrefDOI} \doi{https://doi.org/10.1101/240614} \end{APACrefDOI}
\PrintBackRefs{\CurrentBib}

\bibitem [\protect \citeauthoryear {%
RichardWebster%
, Anthony%
\BCBL {}\ \BBA {} Scheirer%
}{%
RichardWebster%
, Anthony%
\BCBL {}\ \BBA {} Scheirer%
}{%
{\protect \APACyear {2018}}%
}]{%
richardwebster2016psyphy}
\APACinsertmetastar {%
richardwebster2016psyphy}%
\begin{APACrefauthors}%
RichardWebster, B.%
, Anthony, S.%
\BCBL {}\ \BBA {} Scheirer, W.%
\end{APACrefauthors}%
\unskip\
\newblock
\APACrefYearMonthDay{2018}{}{}.
\newblock
{\BBOQ}\APACrefatitle {PsyPhy: A Psychophysics Driven Evaluation Framework for
  Visual Recognition} {Psyphy: A psychophysics driven evaluation framework for
  visual recognition}.{\BBCQ}
\newblock
\APACjournalVolNumPages{IEEE Transactions on Pattern Analysis and Machine
  Intelligence}{}{}{1-1}.
\newblock
\begin{APACrefDOI} \doi{10.1109/TPAMI.2018.2849989} \end{APACrefDOI}
\PrintBackRefs{\CurrentBib}

\bibitem [\protect \citeauthoryear {%
RichardWebster%
, Kwon%
, Clarizio%
, Anthony%
\BCBL {}\ \BBA {} Scheirer%
}{%
RichardWebster%
, Kwon%
\BCBL {}\ \protect \BOthers {.}}{%
{\protect \APACyear {2018}}%
{\protect \APACexlab {{\protect \BCnt {1}}}}}]{%
osti-10144516}
\APACinsertmetastar {%
osti-10144516}%
\begin{APACrefauthors}%
RichardWebster, B.%
, Kwon, S\BPBI Y.%
, Clarizio, C.%
, Anthony, S\BPBI E.%
\BCBL {}\ \BBA {} Scheirer, W\BPBI J.%
\end{APACrefauthors}%
\unskip\
\newblock
\APACrefYearMonthDay{2018{\protect \BCnt {1}}}{}{}.
\newblock
{\BBOQ}\APACrefatitle {Visual Psychophysics for Making Face Recognition
  Algorithms More Explainable} {Visual psychophysics for making face
  recognition algorithms more explainable}.{\BBCQ}
\newblock
\APACjournalVolNumPages{European Conference on Computer Vision 2018}{15}{}{}.
\newblock
\begin{APACrefURL} \url{https://par.nsf.gov/biblio/10144516} \end{APACrefURL}
\newblock
\begin{APACrefDOI} \doi{10.1007/978-3-030-01267-0} \end{APACrefDOI}
\PrintBackRefs{\CurrentBib}

\bibitem [\protect \citeauthoryear {%
RichardWebster%
, Kwon%
, Clarizio%
, Anthony%
\BCBL {}\ \BBA {} Scheirer%
}{%
RichardWebster%
, Kwon%
\BCBL {}\ \protect \BOthers {.}}{%
{\protect \APACyear {2018}}%
{\protect \APACexlab {{\protect \BCnt {2}}}}}]{%
RichardWebster-2018-ECCV}
\APACinsertmetastar {%
RichardWebster-2018-ECCV}%
\begin{APACrefauthors}%
RichardWebster, B.%
, Kwon, S\BPBI Y.%
, Clarizio, C.%
, Anthony, S\BPBI E.%
\BCBL {}\ \BBA {} Scheirer, W\BPBI J.%
\end{APACrefauthors}%
\unskip\
\newblock
\APACrefYearMonthDay{2018{\protect \BCnt {2}}}{September}{}.
\newblock
{\BBOQ}\APACrefatitle {Visual Psychophysics for Making Face Recognition
  Algorithms More Explainable} {Visual psychophysics for making face
  recognition algorithms more explainable}.{\BBCQ}
\newblock
\BIn{} \APACrefbtitle {Proceedings of the European Conference on Computer
  Vision (ECCV).} {Proceedings of the european conference on computer vision
  (eccv).}
\PrintBackRefs{\CurrentBib}

\bibitem [\protect \citeauthoryear {%
Riesenhuber%
\ \BBA {} Poggio%
}{%
Riesenhuber%
\ \BBA {} Poggio%
}{%
{\protect \APACyear {1999}}%
}]{%
riesenhuber1999hierarchical}
\APACinsertmetastar {%
riesenhuber1999hierarchical}%
\begin{APACrefauthors}%
Riesenhuber, M.%
\BCBT {}\ \BBA {} Poggio, T.%
\end{APACrefauthors}%
\unskip\
\newblock
\APACrefYearMonthDay{1999}{}{}.
\newblock
{\BBOQ}\APACrefatitle {Hierarchical models of object recognition in cortex}
  {Hierarchical models of object recognition in cortex}.{\BBCQ}
\newblock
\APACjournalVolNumPages{Nature neuroscience}{2}{11}{1019--1025}.
\PrintBackRefs{\CurrentBib}

\bibitem [\protect \citeauthoryear {%
Russakovsky%
\ \protect \BOthers {.}}{%
Russakovsky%
\ \protect \BOthers {.}}{%
{\protect \APACyear {2015}}%
}]{%
ILSVRC15}
\APACinsertmetastar {%
ILSVRC15}%
\begin{APACrefauthors}%
Russakovsky, O.%
, Deng, J.%
, Su, H.%
, Krause, J.%
, Satheesh, S.%
, Ma, S.%
\BDBL {}Fei-Fei, L.%
\end{APACrefauthors}%
\unskip\
\newblock
\APACrefYearMonthDay{2015}{}{}.
\newblock
{\BBOQ}\APACrefatitle {{ImageNet Large Scale Visual Recognition Challenge}}
  {{ImageNet Large Scale Visual Recognition Challenge}}.{\BBCQ}
\newblock
\APACjournalVolNumPages{International Journal of Computer Vision
  (IJCV)}{115}{3}{211-252}.
\newblock
\begin{APACrefDOI} \doi{10.1007/s11263-015-0816-y} \end{APACrefDOI}
\PrintBackRefs{\CurrentBib}

\bibitem [\protect \citeauthoryear {%
Schrimpf%
\ \protect \BOthers {.}}{%
Schrimpf%
\ \protect \BOthers {.}}{%
{\protect \APACyear {2020}}%
}]{%
schrimpf2020integrative}
\APACinsertmetastar {%
schrimpf2020integrative}%
\begin{APACrefauthors}%
Schrimpf, M.%
, Kubilius, J.%
, Lee, M\BPBI J.%
, Murty, N\BPBI A\BPBI R.%
, Ajemian, R.%
\BCBL {}\ \BBA {} DiCarlo, J\BPBI J.%
\end{APACrefauthors}%
\unskip\
\newblock
\APACrefYearMonthDay{2020}{}{}.
\newblock
{\BBOQ}\APACrefatitle {Integrative benchmarking to advance neurally mechanistic
  models of human intelligence} {Integrative benchmarking to advance neurally
  mechanistic models of human intelligence}.{\BBCQ}
\newblock
\APACjournalVolNumPages{Neuron}{}{}{}.
\PrintBackRefs{\CurrentBib}

\bibitem [\protect \citeauthoryear {%
Simonyan%
\ \BBA {} Zisserman%
}{%
Simonyan%
\ \BBA {} Zisserman%
}{%
{\protect \APACyear {2015}}%
}]{%
simonyan2015deep}
\APACinsertmetastar {%
simonyan2015deep}%
\begin{APACrefauthors}%
Simonyan, K.%
\BCBT {}\ \BBA {} Zisserman, A.%
\end{APACrefauthors}%
\unskip\
\newblock
\APACrefYearMonthDay{2015}{}{}.
\newblock
\APACrefbtitle {Very Deep Convolutional Networks for Large-Scale Image
  Recognition.} {Very deep convolutional networks for large-scale image
  recognition.}
\PrintBackRefs{\CurrentBib}

\bibitem [\protect \citeauthoryear {%
{Szegedy}%
\ \protect \BOthers {.}}{%
{Szegedy}%
\ \protect \BOthers {.}}{%
{\protect \APACyear {2015}}%
}]{%
7298594}
\APACinsertmetastar {%
7298594}%
\begin{APACrefauthors}%
{Szegedy}, C.%
, {Wei Liu}%
, {Yangqing Jia}%
, {Sermanet}, P.%
, {Reed}, S.%
, {Anguelov}, D.%
\BDBL {}{Rabinovich}, A.%
\end{APACrefauthors}%
\unskip\
\newblock
\APACrefYearMonthDay{2015}{}{}.
\newblock
{\BBOQ}\APACrefatitle {Going deeper with convolutions} {Going deeper with
  convolutions}.{\BBCQ}
\newblock
\BIn{} \APACrefbtitle {2015 IEEE Conference on Computer Vision and Pattern
  Recognition (CVPR)} {2015 ieee conference on computer vision and pattern
  recognition (cvpr)}\ (\BPG~1-9).
\newblock
\begin{APACrefDOI} \doi{10.1109/CVPR.2015.7298594} \end{APACrefDOI}
\PrintBackRefs{\CurrentBib}

\end{thebibliography}

\end{document}